# DPOCL: A Principled Approach to Discourse Planning

R. Michael Young
Intelligent Systems Program
University of Pittsburgh
Pittsburgh, PA, 15260
myoung+@pitt.edu

Johanna D. Moore
Department of Computer Science and
Learning Research and Development Center
University of Pittsburgh
Pittsburgh, PA 15260
jmoore@cs.pitt.edu

**Abstract**

Research in discourse processing has identified two representational requirements for discourse planning systems. First, discourse plans must adequately represent the intentional structure of the utterances they produce in order to enable a computational discourse agent to respond effectively to communicative failures [15]. Second, discourse plans must represent the informational structure of utterances. In addition to these representational requirements, we argue that discourse planners should be formally characterizable in terms of soundness and completeness.

## 1 Introduction

Research in discourse processing has identified two representational requirements for discourse planning systems. First, discourse plans must adequately represent the intentional structure of the utterances they produce in order to enable a computational discourse agent to respond effectively to communicative failures [15]. Second, discourse plans must represent the informational structure of utterances. Discourse interpretation requires that an agent be able to recognize the relationships between the information conveyed in consecutive elements of discourse (e.g., [7, 16]). Choosing syntactic structures and connective markers that convey these relationships requires that a discourse generator represent informational [19, 21, 22] as well as intentional [4] structure. Because there is not a fixed, one-to-one mapping between intentional and informational structures, discourse plans must include an explicit representation of both types of structure [15, 16].

In addition to these representational requirements, we argue that discourse planners should meet certain computational requirements. Most current discourse planners are based on the original NOAH [20] model of hierarchical planning [1, 2, 9, 13, 15]. These systems rely on customized planning algorithms with procedural semantics for the purposes of solving specific text-planning problems. The informal construction of these systems and their application to particular problems have resulted in successful text generation for specific domains and text types. However, careful analysis of these programs shows that there is nothing in their semantics to prevent them from generating incorrect plans, generating plans with redundant steps, or failing to find plans in situations where they exist. To the extent that these planners have been able to avoid these problems, they have done so by severely limiting the expressive power of action descriptions and/or requiring the designer to handcraft each action description to fit correctly into the *ad hoc* semantics of the specific plan for which the action is intended. As the number of operators for such systems increases it becomes impractical for their designers to maintain their consistency.

To overcome these limitations, we argue that a discourse planning algorithm should be formally sound and complete (or at least be formally characterizable in terms of these properties). While these formal characteristics may need to be relaxed in order to construct a planner for any given application, it is important to determine exactly how a planning algorithm fails to meet these requirements. Without such a characterization one cannot specify what class of plans, and thus what class of discourses, are generated by a discourse planning system.

In this paper we provide a general characterization of previous discourse planning systems in terms of these properties. We then describe a new discourse planning algorithm that extends recent work on partial-order, causal link (POCL) planning systems to represent hierarchical discourse plans. We show how this algorithm, called DPOCL (Decompositional POCL), provides a formal and explicit model of intentional and informational structure in its plans. In addition, we discuss DPOCL's formal properties.

## 2 Representation in Discourse Plans

Previous approaches have viewed the discourse planner as a means to producing a specification of a discourse that can be given to a text realization system in order to produce a series of sentences in a natural language. Recent work has shown that plans play a much larger role in agent interaction [18]. In particular, the structure of discourse plans plays a role in the comprehension of the discourse [6, 11, 16] and contributes to the nature of subsequent communication [15, 24].

### 2.1 Representing Intentional Structure

As has been noted [15, 16, 24], a precise definition of intention in discourse plans is crucial for enabling systems to respond appropriately to failures of their communicative actions. When a hearer reveals that an in-tended effect of a previous discourse did not succeed, the speaker should re-try to achieve that effect. If, however, the effect that failed was not an *intended* effect, the speaker need not generate an alternative response to achieve it. Alternatively, if the effect that failed was intended, but served only as a precondition of an action whose intended effects succeeded despite the failure, then again the speaker may chose not to respond. Clearly, differentiating between intended and unintended effects of discourse actions is critical for generating appropriate responses.

As Maybury has pointed out [13], a realistic description of communicative action requires a representation that allows individual actions to have more than one effect on the mental state of the hearer. In particular, abstract communicative actions need to be described in a way that represents at least some of the effects of the steps in their subplans. Allowing action descriptions that have multiple effects complicates the definition of intentional structure. The reason for inserting a step in a plan is to establish some intended condition(s). However, when steps have multiple effects, it may be the case that only some of these effects are used to establish intended conditions in any given context. Any effects of a step that do not play a role in establishing such conditions in a given plan are considered side effects.

In Section 6.1, we formally define intention in the DPOCL framework, and show how intended effects are distinguished from side effects when action descriptions may have multiple effects.

## 2.2 Representing Informational Structure

Just as the structure of a discourse reflects the intentions of the speaker, that structure also reflects the way in which domain content is used to achieve intended effects. This informational structure captures relationships that hold between objects in the domain of discourse. In an explanation, for example, one utterance may describe an event that can be presumed to be the cause of another event described in the subsequent utterance.

Clearly intention and information are closely related. An important component of an agent's linguistic capability is the knowledge of what types of information can be used to achieve communicative intentions. Hearers may be able to determine what the speaker is trying to do because of what the hearer knows about the world or what she knows about what the speaker believes about the world. Alternatively, the hearer may be able to figure out what the speaker believes about the world by recognizing what the speaker is trying to do in the discourse [16]. In Section 6.1, we describe how decomposition operators in DPOCL capture the relationship between intentional and informational structure.

## 3 Desiderata for Planning Algorithms

A formal characterization of the capabilities of discourse planning algorithms is essential to understanding their limitations and is necessary before one can make claims about the kinds of discourse plans those planners can produce.

## 3.1 Completeness

The planning process of a generative planner is typically viewed as a search through the space of possible plans to locate a solution for a given planning problem. For some planning problems, no solutions exist. For others, many solutions can be constructed. A general-purpose discourse planner cannot anticipate the structure of the solutions to every problem. In order for these planners to be useful, they must be able to construct all solutions. Planners that are guaranteed to find all solutions to a planning problem are *complete*.

Suppose there is a class of solutions to a planning problem that a discourse planning algorithm cannot find. It may be the case that the most appropriate solutions to the problem fall entirely into this class. If this happens, the planner will only be able to construct the less-desirable plans. It may also be the case that the *only* solutions to a planning problem fall into this class. If this happens, the planner will unnecessarily report failure. Consider those discourses in which individual utterances play several roles. Maier [12] describes the need for a system to generate this type of discourse and Hobbs [7] provides an example of one such discourse. We provide another example here:

> Lucentio has asked for Bianca's hand. He always considered her Senior Baptista's fairest daughter. That is also why Lucentio always chose her to model for his paintings.

Here the second sentence provides support for the hearer's acceptance of both surrounding sentences.

In a discourse planning model, these multi-role utterances correspond to actions that are part of subplans for two different parent actions. That is, the plans that represent these actions are structured as directed acyclic graphs (DAGS) rather than trees. Planning algorithms that are incomplete because they can only produce tree-structured plans are not able to generate plans for this class of discourse. For the example above, these types of planners would produce less appropriate plans where the second sentence appeared twice as support in two distinct subtrees.

## 3.2 Soundness

Any system that plans before it acts assumes that its model of action is a useful one. Given that a system is using such a model, the plans that it produces should at least be internally consistent. That is, these plans should not have steps that interfere with one another. Furthermore, the planner should continue to add steps to a plan until the model indicates that all the plan's goals have been accounted for. Planning algorithms that have these properties are called *sound*.

Given a model of planning where actions are related both causally and decompositionally, a sound planning algorithm must consider two factors when constructing plans. First, for every step in a plan, the planner must ensure that each precondition of that step will be true just prior to its execution [3]. Second, the planner must consider the manner in which the steps of a subplan achieve the goals of the parent [23, 25]. While a parent step specifies the effects it has on the mental state of the hearer, it is the responsibility of the executable steps at the leaves of the subplan rooted at the parent step to ensure that those conditions are indeed established.

Note that the soundness of a planning algorithm does not guarantee the success of the plans it produces. However, the information about causal and decompositional relationships recorded in sound plans is crucial for determining where the planning model is in error and how to replan when an execution failure occurs.

## 4 Previous Discourse Planning Systems

Most current discourse planners (e.g., [1, 2, 9, 13, 15]) are based on the original NOAH model of hierarchical planning. They rely on customized planning algorithms with procedural semantics for the purposes of solving specific text-planning problems, and thus their representations and algorithms suffer from being unprincipled and difficult to analyze. Although these systems have resulted in successful text generation for specific domains and text types, careful analysis of these programs shows that there is nothing in their semantics to prevent them from generating incorrect plans, generating plans with redundant steps, or failing to find plans in situations where they exist.

As Hovy et al. [8] point out, these problems stem from an approach to discourse planning that does not clearly distinguish between the representation of communicative action and the design of a planning algorithm that manipulates that representation. In most previous work, there has been no clear separation between the knowledge about the preconditions and effects of communicative acts and the knowledge about planning used to construct discourse plans. To the extent that these planners have been able to avoid generating incorrect or redundant plans, they have done so by severely limiting the expressive power of action descriptions and/or requiring the designer of action descriptions to handcraft each description to fit correctly into the *ad hoc* semantics of the specific plan for which the action is intended. As Hovy et al. describe, when the number of operators for such systems increases, it becomes impractical for their designers to maintain their consistency.

### 4.1 Representation of Discourse Plans

Plans produced by most previous discourse planners have not adequately represented both the causal and decompositional relations between actions in a discourse plan. As a result, their plans do not represent the manner in which preconditions are established, and, in cases where they represent action decomposition, the plans do not capture the relationship between the effects of actions in a subplan and the effects of their parent action. Furthermore, they do not represent intentional and in-formational structure in a way that clearly distinguishes the two. As a result, the intentional and informational structures in their plans are difficult to analyze. Furthermore, the discourse operators for these systems lack the generality that would come from separating the two structures.

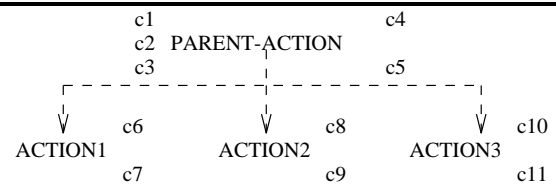

Figure 1: Schematic Discourse Plan Illustrating Parent/Subplan Effects

**Intention** To illustrate these problems, consider the discourse plan shown schematically in Figure 1.[1] This plan has a structure that is typical of those produced by most previous discourse planning systems [2, 9, 13, 15]. In this plan there is no explicit connection between the effects established by the parent action ($c4$ and $c5$) and those established by its subplan ($c6$, $c7$, $c9$, $c10$, $c12$ and $c13$). Previous approaches only represent the relationship between *actions* at different levels; they fail to capture the relationship between the *effects* of those actions. In Figure 1, the top-level goal is $c4 \wedge c5$. Suppose that $c6$ unifies with $c4$, and that $c9$, $c10$, and $c12$ together have a consequence that unifies with $c5$. In this case, $c7$ and $c13$ are side effects of choosing the decomposition of the PARENT-ACTION into ACTION1, ACTION2 and ACTION3. This fact, however, is not captured in the discourse plan of Figure 1. A system relying on this plan could not distinguish intended effects from side effects, and so would be unable to determine that a different response is warranted when $c6$ fails than when $c7$ fails.

In addition, there is no explicit representation of the relationship between two steps when one establishes a precondition for another. In Figure 1, ACTION3 has $c11$ as a precondition. Suppose that both $c10$ and $c7$ unify with $c11$. If $c10$ fails it is possible that $c7$ will serve to establish the condition needed by $c11$. Without a representation of the causal roles that these effects play, a system cannot determine whether an additional response is required.

**Informational Structure in Previous Systems** Most previous planning systems do not provide an explicit representation for either intentional or informational structure. As noted in Hovy, et al [8], to the extent that informational constraints were represented, each set of constraints was duplicated for many similar discourse operators. Many of these operators differed only in their intentional structure. As described in [16], combining intentional and informational representations in this way can result in a proliferation of operators. Every intentional structure must be paired with every informational one, possibly requiring as many as $n \times m$ operators for domains with $n$ intentional and $m$ informational structures.

---
[1] In this plan, the dashed arcs indicate the decomposition of PARENT-ACTION into the actions in its subplan. The $c_i$'s represent conditions in the world – those to the left of an action are the action's preconditions and those to the right of an action are its effects.

## 4.2 Computational Properties

While previous discourse planners have been successful at generating appropriately structured plans for specific domains, these systems have ignored the analysis of the formal properties of the planning algorithms that produce them. As has been noted in [20, 3], NOAH, and consequently those discourse planners based on it, use *ad hoc* procedures for the construction of plans. As a result, the formal properties of these planning algorithms are difficult to characterize. While a complete analysis of the planning algorithms used by previous systems [2, 9, 13, 15] is beyond the scope of this paper, several properties of these algorithms are straightforward to describe.

First, these planners do not guarantee that a step's preconditions hold prior to the step's execution and thus they are not sound. Furthermore, there is no relationship in any of these planners between the effects of parent actions and their subplans – planning to achieve an effect at one level of abstraction does not guarantee that the effect is realized by any combination of executable actions.

Second, these planners are not complete. While there may be many classes of plans that these systems cannot generate, their incompleteness can easily be seen when considering two factors. First, all of these systems use tree-structured plan representations. As a result, they cannot produce discourse plans where individual components play a role in more than one subplan. Second, most current discourse planning systems restrict steps in subplans to be totally ordered with respect to one another. For total-order planners to be complete they must be able to construct every possible step ordering.

While the sacrifice of formal properties may be necessary for constructing an efficient implementation, it is important to characterize the conditions under which a planning system falls short of soundness or completeness. By characterizing the soundness of a planner two things become apparent. First, the conditions under which a planner will introduce flaws into a plan are completely characterized. Second, the nature of the flaws that might be introduced under those conditions are specified. Similarly, characterizing the completeness of a planner specifies the classes of plans that can and cannot be produced by a planner. Without an understanding of these properties for a given algorithm, it is impossible fully to evaluate its usefulness for a particular application.

## 5 The DPOCL Discourse Planner

The DPOCL discourse planner is an extension to recent partial-order causal link planners [14, 17]. In POCL planners, a plan is represented as a set of partially-ordered steps connected by causal links. Two steps in a plan are connected by a causal link when the effect of the first step is used to establish the precondition of the second step. Steps and corresponding links are added to the plan to establish unsatisfied preconditions, and additional constraints are placed on the plan only when needed to maintain consistency. Previous POCL planners have been non-hierarchical; DPOCL provides an extension that introduces action decomposition into the POCL framework. For a complete definition of DPOCL see [25].

In the following discussion we will refer to the sample discourse from Section 3.1. Figure 2 shows an example of a DPOCL plan structure for this discourse. Consider the subplan for Support(modeled(L,B)), rooted at the step marked as step #1.[2] A decomposition link (shown using dashed arcs) connects Support(modeled(L,B)) to the begin and end-subplan steps bounding its subplan.[3] This subplan is made up of the two Cause-to-Believe steps and the Combine-Belief step shown in between the begin-subplan and end-subplan. A causal link (shown using a solid arc and labeled with the effect that it contributes) connects Cause-to-Believe(fairest(L,B)) to the End-Subplan step.

The manner in which a hearer combines the information in an utterance with his prior beliefs is critical to the generation of the utterance. Most previous work has made use of highly simple models of this process: for instance, it has assumed that the effect of asserting a proposition $p$ is that the hearer believes $p$. In fact, a speaker may go to great lengths to convince the hearer of the truth of a proposition. She may first assert it, then support it, and then provide support for the intermediate statement. In such a case, the speaker presumably believes that the combination of utterances is what leads the hearer to accept the main proposition.

This phenomenon is represented by the Combine-Belief($\vec{x}$) action, where $\vec{x}$ is a vector of relevant beliefs. This Combine-Belief action provides an abstract model for an action taken by the hearer rather than by the speaker. A complete model of the manner in which a user combines belief from several utterances is beyond the scope of this paper. Further formal work in this area, such as that in [10] is essential for accurately representing the structure of discourse plans. DPOCL provides a

---

[2]Subplans in this figure are grouped inside rounded boxes for ease of reference.

[3]DPOCL uses the standard POCL technique of encoding the initial conditions and the goals of a planning problem as the effects of a null initial action and the preconditions of a null final action, respectively. Similarly, every subplan is bounded by a null start-subplan and a null end-subplan. Each start-subplan has as its effects the preconditions of its parent action, and each end-subplan has as its preconditions the effects of its parent action.

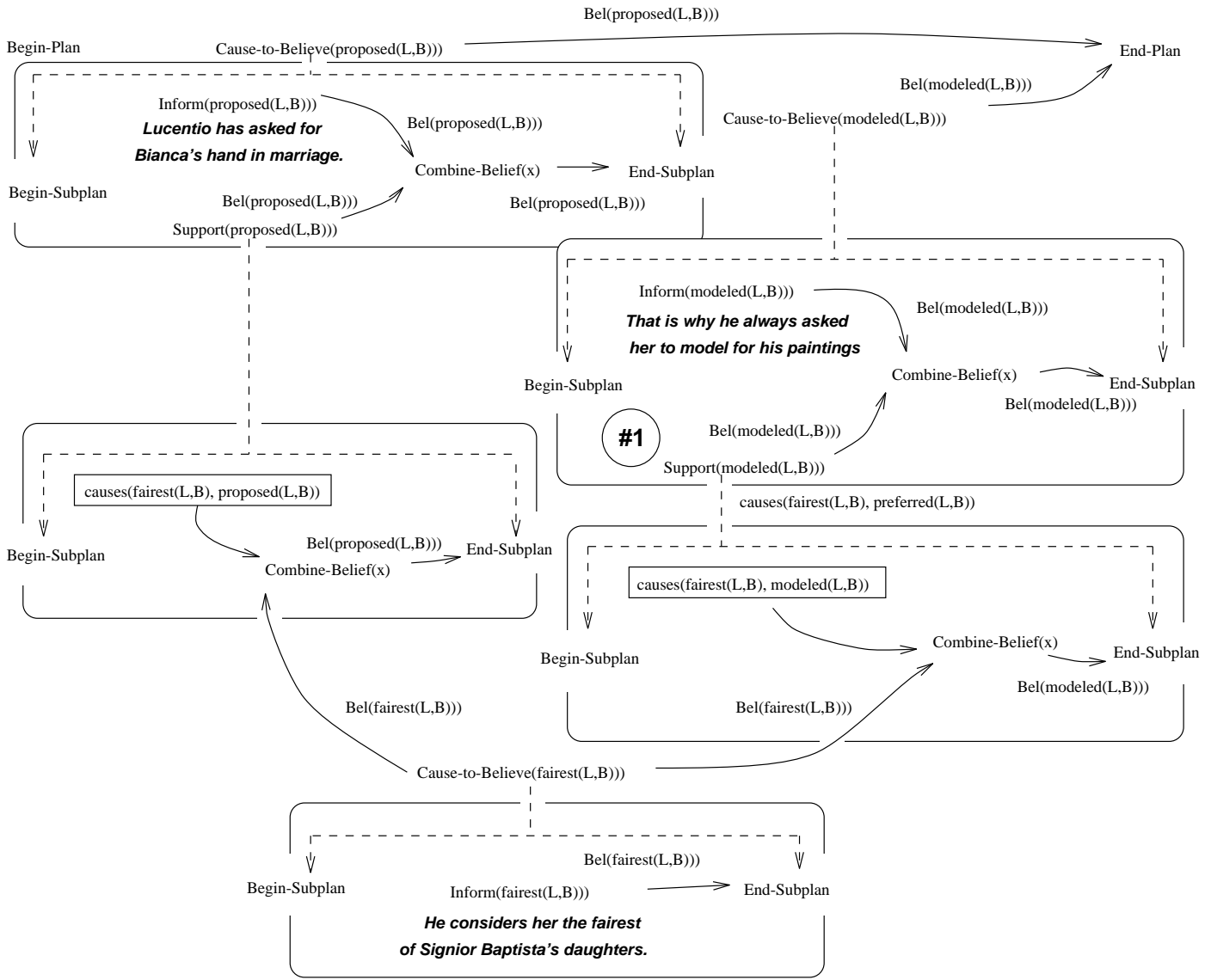

Figure 2: A Complete Discourse Plan

### 5.1 Representation in DPOCL

The representation of each action in DPOCL is separated into two parts corresponding to the causal and decompositional roles the action plays: the action operator, and a set of decomposition operators. The action operator captures the action's preconditions and effects, sets of first-order unquantified sentences similar to the typical precondition and add/delete lists of STRIPS [5]. Each decomposition operator represents a single-layer expansion of a composite step, essentially providing a partial specification for the subplan that achieves the parent step's effects given its preconditions. In addition to specifying the steps in the subplan, the decomposition operator specifies any variable binding and temporal ordering constraints between the steps, and the causal links between steps of the subplan that enable them to establish the parent step's effects.

Figure 3 shows the action operator and one decomposition operator for the Support act.[4] As we see in the action operator in this figure, Support(?*prop*) has the effect of increasing the belief in proposition ?*prop* for the hearer. The decomposition operator in Figure 3 was responsible for expanding the Support labeled #1 in

---

[4]The operators in this figure are shown with some detail omitted for clarity.

```
Action
Header:           Support(?prop)
Preconditions:    not(Believe(?prop))
Effects:          Bel(?prop)
Bindings:         none

Decomposition
Header:           Support(?prop1)
Constraints:      causes(?prop2, ?prop1)
Steps:            Start, Cause-to-Believe-1(?prop2)
                  Cause-to-Believe-2(causes(?prop2,?prop1))
                  Combine-Belief, Final
Links:            <Combine-Belief, ?prop, ?prop, Final>
Bindings:         none shown
Orderings:        none shown
```

Figure 3: Support Action and Decomposition Operators

Figure 2. The subplan specified by this decomposition has three steps in its body: two Cause-to-Believe actions and a combination of belief by the hearer to strengthen her belief in ?prop. The constraints placed on this decomposition restrict the propositions used in the Cause-to-Believe steps to be ones that cause the proposition being supported. Decompositional constraints are discussed further in Section 6.1

This decomposition operator is only a partial specification of the subplan for the Support step. In DPOCL, when a subplan is only partially specified, the planner is free to complete the subplan by using steps already appearing in the plan. In this way, DPOCL can avoid generating plans with redundant communicative actions.

## 5.2 Overview of the DPOCL Algorithm

In DPOCL, the process of creating a completed plan involves iterating through a loop that chooses between refining the current plan decompositionally or refining the plan causally and then modifying the plan to ensure that the refinement has not introduced any errors. Figure 4 summarizes the DPOCL planning algorithm. For a complete definition, see [25].

Causal refinement in DPOCL is essentially identical to causal refinement in previous POCL planners. An unsatisfied precondition of some step in the plan is selected and a causal link is added to establish the needed condition. Decompositional refinement essentially creates a subplan for some composite action and adds the subplan to the plan. First, a decomposition operator for the chosen step is selected and the steps indicated in the operator are added to the plan. These steps are created in one of two ways. In the first case, a step is created by selecting an action operator of the correct action type and instantiating a new step just as is done when a new step is added during causal refinement. In the second case, a step is added to the subplan by finding a step of the correct action type that already exists in the plan and using that step in the appropriate place in the new subplan.

The DPOCL algorithm ensures that a subplan's actions establish the effects of the parent action in a straightforward manner. The preconditions of a sub-plan's final step are an copy of the effects of the subplan's parent step. The DPOCL planner will attempt to achieve them through causal refinement just as it achieves all other unsatisfied preconditions. In this way we guarantee that the effects of every composite action are achieved by the steps in its subplan. Furthermore, the exact relationship between the actions in a subplan and the establishment of those effects is made explicit in the causal links establishing those conditions in the subplan.

As a result of adding steps to a plan, newly created steps may introduce threats to existing causal links. A step, $S_a$, *threatens* a causal link between two steps $S_b$ and $S_c$ when $S_a$ might occur between $S_b$ and $S_c$ and one of $S_a$'s effects might undo the condition established in the causal link. To ensure that no causal links are undone by plan refinement, each threat in a plan is eliminated before planning proceeds. This is done either by ordering the steps so that the threatening step cannot occur between the two causally-linked steps or by restricting the variable bindings of the steps to eliminate harmful interactions.

## 6 DPOCL's Properties

Plan structures in DPOCL represent three critical components. First, every causal connection between some

---

**Termination:** If the plan is inconsistent, then backtrack. Otherwise, remove unused step and return the plan.

**Plan Refinement:** Non-deterministically do one of the following:

1. **Causal Planning:**
   (a) **Goal Selection:** Nondeterministically select a goal.
   (b) **Operator Selection:** Add a step to the plan that adds an effect that can be unified with the goal (either by instantiating the step from the operator library or by finding a step already in the plan). If no such step exists, backtrack. Otherwise, add the binding constraints required for the conditions to unify, an ordering constraint that orders the new step before the goal step and add the causal link between the two.

2. **Decompositional Planning:**
   (a) **Action Selection:** Nondeterministically select some unexpanded composite step in the plan.
   (b) **Decomposition Selection:** Nondeterministically chose an appropriate decomposition schema for this action whose constraints are satisfied. Add the steps and subplan components of the decomposition schema to the plan and update the list of decomposition links to indicate the new subplan.

**Threat Resolution:** Find any step that might threaten to undo any causal link. For every such step, nondeterministically do one of the following:

- **Promotion:** If possible, move the threatened steps to occur before the threat in the plan.
- **Demotion:** If possible, move the threatened steps to occur after the threat in the plan.
- **Separation:** If possible, add binding constraints on the steps involved so that no conflict can arise.

**Recursive Invocation:** Call the planner recursively with the new plan structure.

Figure 4: DPOCL Planning Algorithm

**Definition 1 (Intended Effect)** *Let s be some step in a plan and $e_s$ be an effect of s. Effect $e_s$ is intended precisely when at least one of the following conditions holds:*

- *There is some causal link from s to the final step of the plan such that $e_s$ establishes one of the goals of the plan.*
- *There is some causal link from s to some step $s_f$ where $s_f$ is the final step of a subplan for a parent action $s_p$ such that*
  - *$e_s$ establishes one of the goals of the subplan (that is, a precondition of $s_f$) and*
  - *the corresponding effect $e_{s_p}$ of $s_p$ is intended.*
- *There is some causal link from s to another step $s'$ such that*
  - *$e_s$ establishes one of the preconditions of $s'$ and*
  - *some effect $e_{s'}$ of $s'$ is intended.*

Figure 5: Intention in DPOCL

step's effect and another step's precondition that relies upon it is marked by a causal link. Second, the connection between the effects of every abstract action and the substeps that achieve those effects are marked by a combination of causal and decompositional links. Finally, the constraints restricting the applicability of decomposition operators are noted for every abstract step expansion. By providing an explicit representation for each of these components an adequate characterization of the intentional and informational structure of the discourse can be made.

## 6.1 DPOCL's Representational Properties

**A Principled Representation of Intention** The formal representation of causal and decompositional connections between steps in the DPOCL plan makes the definition of intention in terms of these concepts straightforward. Informally, an effect is intended if it plays a causal role in the plan. That is, if it is used in a causal link and the step that asserts that effect is connected by that causal link through subsequent causal and decompositional links ultimately to the final step of the plan. The formal definition of an intended effect is shown in Figure 5.

Although the plan shown in Figure 2 does not explicitly illustrate how our representation addresses cases where action descriptions have multiple effects and so distinguishes between intended and side-effects in the same action, our model handles these cases appropriately. Our solution rests on the fact that our model makes a clear distinction between effects of discourse actions that play a role in achieving the top-level goals of the discourse plan and effects that are not causally linked in a way that contributes to the agent's ultimate goals.

**An Explicit Representation of Informational Structure** Decomposition operators in DPOCL enable us to represent the knowledge speakers have about how to use domain information to achieve communicative intentions. For example, one way for a speaker to increase a hearer's belief in a proposition (i.e., to support a proposition) is to describe a plausible cause of that proposition. In DPOCL, we represent this "rule of language" using a decomposition operator as illustrated by the decomposition operator in Figure 3. This operator says that one way to support a proposition *?prop*1 is to find another proposition, *?prop*2, such that causes(*?prop*2, *?prop*1) is true in the domain. If such a *?prop*2 can be found, then the speaker can support *?prop*1 by making the hearer believe *?prop*2 and the relation causes(*?prop*2, *?prop*1) In this way, information in the domain acts to constrain what language rules are appropriate and, given any particular rule, what objects can be referred to when it is used.

The representation of the informational structure in a DPOCL plan is straightforward. Each decomposition operator in DPOCL lists the informational constraints that must hold in order for an abstract action to be achieved by the subplan defined in that operator. During plan generation, informational constraints are checked for consistency whenever a modification is made to the plan and backtracking occurs when a constraint is violated. In addition, these constraints are explicitly recorded in the plan data structure. The informational structure is made available to the realization component that is responsible for transforming the discourse plan into a series of natural language utterances.

## 6.2 DPOCL's Computational Properties

Because DPOCL is built upon well-understood POCL planning algorithms, DPOCL inherits many of these algorithms' formal properties. Specifically, DPOCL is both sound and, for certain classes of plans, complete. Proofs of soundness and completeness can be found in [25]. With respect to the class of plans that DPOCL can generate, DPOCL is *primitive* complete. That is, it can generate all possible sequences of executable actions, but not necessarily all hierarchical structures that could account for those executable actions. In particular, DPOCL cannot generate plans where two abstract steps are ordered one before the other in order to avoid a harmful interaction but some interleaving of the steps in their subplans exists that avoids this interaction. For a more complete description of this restriction on DPOCL's completeness, see [25].

## 7 Discussion

As others have pointed out, the precise representation of intentional and informational structure is critical to the effective use of discourse plans. In addition, we have argued that a formal characterization of the planners that produce those plans is essential to evaluating their usefulness for any given domain. As we discussed in Section 4, while previous work addressed some of these issues, their approaches did not resolve the problems we have identified.

In contrast, the DPOCL planner provides an explicit and formal representation of the intentional and informational structures in its discourse plans. This model clearly differentiates between intended and unintended effects, allowing appropriate responses to discourse failure. In addition, the information constraining each decomposition is formally represented as constraints on the applicability of the decomposition operator. The representation of these constraints is independent from any particular intentional structure formed by the subplan they constrain.

Furthermore, the DPOCL planner builds upon a clear and precise formalism that allows the algorithm to be completely characterized. Specifically, DPOCL is sound and, for some class of plans, complete. It is precisely this formal analysis that allows us to specify exactly what class of plans DPOCL cannot generate. This analysis has not been performed for previous discourse systems and so they cannot similarly characterize their algorithms.

## 8  Acknowledgements

The authors would like to thank the anonymous reviewers for their helpful comments. The research described in this paper was supported by the Office of Naval Research Cognitive and Neural Sciences Division (Grant Number: N00014-91-J-1694). Young is supported by a grant from ONR under the FY93 Augmentation of Awards for Science and Engineering Research Training (ASSERT) Program.

## References

[1] Douglas E. Appelt. *Planning English Sentences*. Cambridge University Press, Cambridge, England, 1985.

[2] Alison Cawsey. *Explanation and Interaction: The Computer Generation of Explanatory Dialogues*. MIT Press, Cambridge, Massachusetts, 1993.

[3] David Chapman. Planning for conjunctive goals. *Artificial Intelligence*, 32, 1987.

[4] Michael Elhadad and Kathleen McKeown. Generating connectives. In *Proceedings of the Thirteenth International Conference on Computational Linguistics*, pages 97–101, Helsinki, 1990.

[5] Richard Fikes and Nils Nilsson. STRIPS: A new approach to the application of theorem proving to problem solving. In James Allen, James Hendler, and Austin Tate, editors, *Readings in Planning*. Morgan Kaufmann, 1990.

[6] Barbara J. Grosz and Candace L. Sidner. Attention, intention, and the structure of discourse. *Computational Linguistics*, 12(3):175–204, 1986.

[7] Jerry Hobbs. On the coherence and structure of discourse. Technical Report CSLI-85-37, Center for the Study of Language and Information, 1985.

[8] Eduard Hovy, Julia Lavid, Elisabeth Maier, Vibhu Mittal, and Cecile Paris. Employing knowledge resources in a new text planner architecture. In *Aspects of Automated Natural Language Generation, Proceedings of the Sixth International Workshop on Natural Language Generation*, Trento, Italy, 1992.

[9] Eduard H. Hovy. Approaches to the planning of coherent text. In Cécile L. Paris, William R. Swartout, and William C. Mann, editors, *Natural Language Generation in Artificial Intelligence and Computational Linguistics*, pages 83–102. Kluwer Academic Publishers, Boston, 1991.

[10] Lynn Lambert. *Recognizing Complex Discourse Acts: A Tripartite Plan-Based Model of Dialogue*. PhD thesis, University of Delaware, 1993.

[11] Lynn Lambert and Sandra Carberry. A tripartite plan-based model of dialog. In *Proceedings, Association for Computational Linguistics*, 1991.

[12] Elisabeth Maier. The extension of a text planner for the treatment of multiple links between text units. In *Proceedings of the 4th European Workshop on Natural Language Generation*, April 1993.

[13] Mark T. Maybury. Communicative acts for explanation generation. *International Journal of Man-Machine Studies*, 37(2):135–172, 1992.

[14] David McAllister and David Rosenblit. Systematic non-linear planning. In *Proceedings of AAAI*, 1991.

[15] Johanna D. Moore and Cécile L. Paris. Planning text for advisory dialogues: Capturing intentional and rhetorical information. *Computational Linguistics*, 19(4):651–695, 1993.

[16] Johanna D. Moore and Martha E. Pollack. A problem for RST: The need for multi-level discourse analysis. *Computational Linguistics*, 18(4):537–544, 1992.

[17] Scott Penberthy and Daniel Weld. UCPOP: A sound, complete partial order planner for adl. In *Proceedings of the Knowledge Representation Conference*, 1991.

[18] Martha E. Pollack. The uses of plans. *Artificial Intelligence*, 57:43–68, 1992.

[19] Dietmar Rosner and Manfred Stede. Customizing RST for the automatic production of technical manuals. In R. Dale, E. Hovy, D. Rosner, and O. Stock, editors, *Proceedings of the Sixth International Workshop on Natural Language Generation*, pages 199–215, Berlin, 1992. Springer-Verlag.

[20] Earl Sacerdoti. *A Structure for Plans and Behavior*. PhD thesis, Stanford University Department of Computer Science, 1975.

[21] Donia Scott and Clarisse Sieckenius de Souza. Getting the message across in RST-based text generation. In R. Dale, C. Mellish, and M. Zock, editors, *Current Research in Natural Language Generation*, pages 47–73. Academic Press, New York, 1990.

[22] Keith Vander Linden, Susanna Cumming, and James Martin. Expressing local rhetorical relations in instructional text. Technical Report 92-43, U. of Colorado, 1992.

[23] Q. Yang. Formalizing planning knowledge for a hierarchical planner. *Computational Intelligence*, 6, 1990.

[24] R. Michael Young, Johanna D. Moore, and Martha E. Pollack. Towards a principled representation for discourse plans. In *Proceedings of the Sixteenth Annual Conference of the Cognitive Science Society*, 1994.

[25] R. Michael Young, Martha E. Pollack, and Johanna D. Moore. Decomposition and causality in partial order planning. In *Proceedings of the Second International Conference on AI and Planning Systems*, 1994.